\documentclass[lettersize,journal]{IEEEtran}

\usepackage{amsmath,amsfonts}
\usepackage{array}
\usepackage[caption=false,font=normalsize,labelfont=sf,textfont=sf]{subfig}
\usepackage{textcomp}
\usepackage{stfloats}
\usepackage{url}
\usepackage{verbatim}
\usepackage{graphicx}
\usepackage{cite}

\usepackage{booktabs}
\usepackage{enumitem}  %
\usepackage{colortbl}  %
\usepackage{xcolor}

\usepackage{multirow}
\usepackage[ruled]{algorithm2e}
\makeatletter
\newcommand{\removelatexerror}{\let\@latex@error\@gobble}
\makeatother

\begin{document}

\title{AVadCLIP: Audio-Visual Collaboration for Robust Video Anomaly Detection}

\author{Peng Wu, Wanshun Su, Guansong Pang~\IEEEmembership{Member, IEEE}, Yujia Sun, Qingsen Yan~\IEEEmembership{Member, IEEE}, \\Peng Wang~\IEEEmembership{Member, IEEE} and Yanning Zhang~\IEEEmembership{Fellow, IEEE}
\thanks{Peng Wu, Wanshun Su, Qingsen Yan, Peng Wang, and Yanning Zhang are with the School of Computer Science, Northwestern Polytechnical University, China. E-mail: \{xdwupeng, suws0616, qingsenyan\}@gmail.com; \{peng.wang, ynzhang\}@nwpu.edu.cn.\\
Guansong Pang is with the School of Computing and Information Systems, Singapore Management University, Singapore. E-mail: pangguansong@gamil.com.\\
Yujia Sun is with the School of Artifical Intelligence, Xidian University, China. E-mail: yjsun@stu.xidian.edu.cn.}
\thanks{Manuscript received April 19, 2021; revised August 16, 2021.}}

\markboth{Journal of \LaTeX\ Class Files,~Vol.~14, No.~8, August~2021}%
{Shell \MakeLowercase{\textit{et al.}}: A Sample Article Using IEEEtran.cls for IEEE Journals}


\maketitle

\begin{abstract}
With the increasing adoption of video anomaly detection in intelligent surveillance domains, conventional visual-only detection approaches often struggle with information insufficiency and high false-positive rates in complex environments. To address these limitations, we present a novel weakly supervised framework that leverages audio-visual collaboration for robust video anomaly detection. Capitalizing on the exceptional cross-modal representation learning capabilities of Contrastive Language-Image Pretraining (CLIP) across visual, audio, and textual domains, our framework introduces two major innovations: an efficient audio-visual fusion that enables adaptive cross-modal integration through lightweight parametric adaptation while maintaining the frozen CLIP backbone, and a novel audio-visual prompt that dynamically enhances text embeddings with key multimodal information based on the semantic correlation between audio-visual features and textual labels, significantly improving CLIP’s generalization for the video anomaly detection task. Moreover, to enhance robustness against modality deficiency during inference, we further develop an uncertainty-driven feature distillation module that synthesizes audio-visual representations from visual-only inputs. This module employs uncertainty modeling based on the diversity of audio-visual features to dynamically emphasize challenging features during the distillation process. Our framework demonstrates superior performance across multiple benchmarks, with audio integration significantly boosting anomaly detection accuracy in various scenarios. Notably, with unimodal data enhanced by uncertainty-driven distillation, our approach consistently outperforms current unimodal VAD methods.
\end{abstract}

\begin{IEEEkeywords}
video anomaly detection, audio-visual collaboration, weakly supervised learning.
\end{IEEEkeywords}

\section{Introduction}
\label{sec:intro}
\IEEEPARstart{V}{ideo} anomaly detection (VAD), as a pivotal technology in intelligent surveillance systems, focuses on identifying anomalous events within videos and has attracted substantial research interest in recent years~\cite{luo2021future, lv2021localizing, wu2021learning, georgescu2021background, zaheer2022stabilizing, cao2024context, liu2024injecting, pu2024learning, wu2024deep, wu2024toward, liu2025crcl}. 
Due to the rarity of anomalies and the high cost of manual annotation, fully supervised frameworks are impractical for large-scale deployment. As a solution, weakly supervised video anomaly detection (WSVAD) methods~\cite{sultani2018real, zaheer2020claws, huang2022weakly, yang2024text} have gained traction, 
aiming to discover latent anomalies under coarse supervision. 
\begin{figure}[t]
  \centering
  \includegraphics[width=0.95\linewidth]{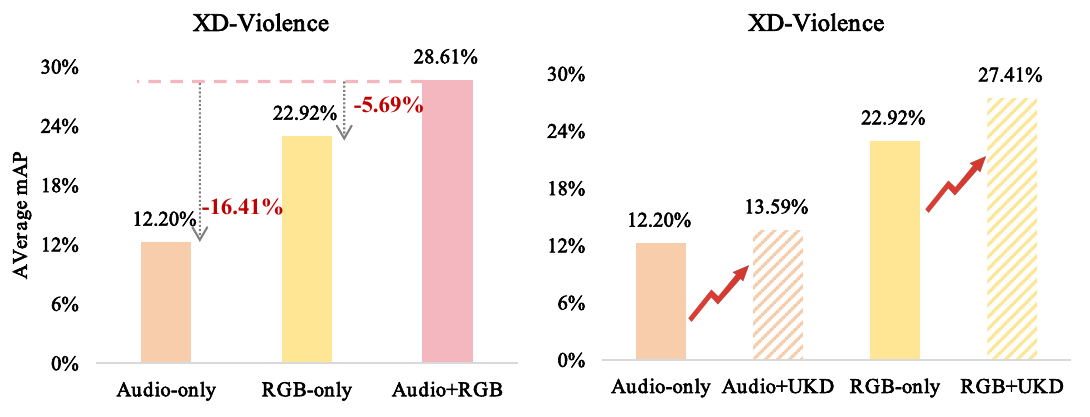}
  \caption{Left: Illustration of audio-visual collaboration effects; Right: Illustration of our proposed distillation (UKD) effects.}
  \label{figure1}
\end{figure}
Current WSVAD methods primarily rely on the multiple instance learning (MIL) framework, using video-level labels for model training~\cite{sultani2018real, zhong2019graph}. Specifically, these approaches treat videos as bags of segments (instances) and distinguish anomalous patterns through the hard attention mechanism (a.k.a Top-K)~\cite{paul2018w}. With the rapid advancement of foundation models, Contrastive Language-Image Pretraining (CLIP)~\cite{radford2021learning} has shown remarkable potential in various downstream tasks, including video understanding~\cite{ju2022prompting, xu2021videoclip}. Building on the remarkable success of CLIP, recent methods like VadCLIP~\cite{wu2024vadclip} and TPWNG~\cite{yang2024text} have advanced WSVAD by leveraging CLIP’s semantic alignment capabilities.

However, these methods, whether CLIP-based or conventional, predominantly rely on unimodal visual information, which often leads to significant detection limitations in complex real-world scenarios. Visual occlusion, extreme lighting variations, and environmental noise can render visual features unreliable or ambiguous~\cite{wu2022weakly, tian2018audio, tian2020unified}. In these challenging conditions, multimodal information, particularly audio, offers indispensable contextual cues that can complement and enhance visual-based detection. For instance, audio remains robust when visual data is compromised, allowing detection of off-camera events. In acoustically rich environments, certain anomalies like explosion, scream, or gunshot exhibit distinct acoustic signatures, making them more discriminative in the audio domain. Similarly, in low-light conditions where visual features degrade, audio serves as a critical supplementary modality. These observations underscore the importance of integrating audio and video modalities, as their complementary nature can significantly enhance the accuracy and robustness of anomaly detection systems in diverse and challenging environments. We illustrate the impact of audio-visual integration for WSVAD in Figure~\ref{figure1}.

Existing attempts~\cite{wu2020not, wu2022weakly, zhou2023dual} to incorporate audio into video anomaly detection typically adopt traditional feature concatenation methods, such as fusing visual features extracted by I3D~\cite{carreira2017quo} or C3D~\cite{tran2015learning} with audio features extracted by VGGish~\cite{gemmeke2017audio}. These approaches fail to fully exploit the potential of multimodal learning, resulting in suboptimal cross-modal integration. Moreover, they overlook the inherent semantic alignment between visual and auditory modalities, which are essential for enhancing anomaly detection performance.

To address these limitations, we propose AVadCLIP, a WSVAD framework that leverages audio-visual collaborative learning to drive audio-visual anomaly detection by CLIP-powered cross-modal alignment. AVadCLIP fully exploits CLIP’s intrinsic capability to establish semantic consistency across vision, text, and audio, ensuring that video anomaly detection is performed within a unified multimodal semantic space rather than merely fusing raw features. Our framework introduces three significant innovations: an efficient audio-visual feature fusion mechanism that is different from the naive feature concatenation and achieves adaptive cross-modal integration through lightweight parametric adaptation while keeping the CLIP backbone frozen; a novel audio-visual prompt mechanism dynamically enriches text label embeddings with key multimodal information, enhancing contextual understanding of videos and enabling more precise identification of different categories; and an uncertainty-driven feature distillation (UKD) module that generates audio-visual-like enhanced features in audio-missing scenarios, ensuring robust anomaly detection performance (as illustrated in Figure~\ref{figure1}). Overall, our AVadCLIP relies on only a small set of trainable parameters, effectively transferring CLIP’s pretrained knowledge to the weakly supervised audio-visual anomaly detection task. Furthermore, by employing a distillation strategy based on data uncertainty modeling, we further transfer the learned knowledge from our audio-visual anomaly detector to a unimodal detector, enabling robust anomaly detection in scenarios with incomplete modalities.

In summary, our main contributions are as follows:
\begin{itemize}
\item
We propose a WSVAD framework that harnesses audio-visual collaborative learning, leveraging CLIP’s multimodal alignment capabilities. By incorporating a lightweight adaptive audio-visual fusion mechanism and integrating audio-visual information through prompt-based learning, our approach effectively achieves CLIP-driven robust anomaly detection in multimodal settings.
\item
We design an uncertainty-driven feature distillation module, which transforms deterministic estimation into probabilistic uncertainty estimation. This enables the model to capture feature distribution variance, ensuring robust anomaly detection performance even with unimodal data.
\item
Extensive experiments on two WSVAD datasets demonstrate that our method achieves superior performance in audio-visual scenarios, while maintaining robust anomaly detection results even in audio-absent conditions. 
\end{itemize}

\section{Related Work}
\label{sec:relatedwork}
\subsection{Video Anomaly Detection}
Video anomaly detection has been extensively studied in recent years, with existing approaches broadly categorized into semi-supervised and weakly supervised methods. Among them, semi-supervised methods primarily rely on normal video clips for training and identify anomalies by detecting deviations from learned normal patterns during inference. These methods commonly adopt self-supervised learning techniques~\cite{huang2022self, shi2023video, huang2024long}, such as reconstruction~\cite{cong2011sparse, luo2017revisit} or prediction~\cite{liu2018future, cao2024scene}. Reconstruction-based methods assume that the model can effectively reconstruct normal videos, whereas abnormal videos, due to distributional discrepancies, result in significant reconstruction errors. Autoencoders~\cite{hasan2016learning, gong2019memorizing} are widely employed to capture normal pattern features, with reconstruction error serving as an anomaly indicator. Prediction-based methods~\cite{yang2023video} utilize models to forecast future frames, detecting anomalies based on prediction errors. However, a key limitation of semi-supervised methods is their tendency to overfit normal patterns, leading to poor generalization to unseen anomalies. 

Weakly supervised methods, in contrast, typically adopt the MIL framework, requiring only video-level anomaly labels and significantly reducing annotation costs. The
classic work, DeepMIL~\cite{sultani2018real}, which employs a ranking loss to distinguish normal from anomalous instances. Furthermore, two-stage self-training strategy has been proposed to further enhance detection, where high-confidence anomalous regions identified during MIL training serve as pseudo-labels for a secondary refinement phase~\cite{li2022self, feng2021mist, cho2023look}. 
With the rise of Vision-Language Models (VLMs)~\cite{chen2023vlp}, CLIP has shown remarkable cross-modal capabilities and is increasingly applied to WSVAD. VadCLIP~\cite{wu2024vadclip}, the first CLIP-based WSVAD method, integrates textual priors via text and visual prompts, enhancing anomaly detection. Building on this, TPWNG~\cite{yang2024text} refines feature learning through a two-stage approach. Recent research trends focus on large model driven strategies, e.g., training-free frameworks~\cite{zanella2024harnessing, yang2024follow}, spatiotemporal anomaly detection~\cite{wu2024weakly}, and open-scene anomaly detection~\cite{wu2024open}. {Recent advances in multi-modal fusion~\cite{ding2025learnable} introduce powerful frameworks combining diverse modalities such as visual and audio features. For instance, AVCL~\cite{meng2025audio} and DSRL~\cite{NEURIPS2024_1f471322} have shown significant promise in improving anomaly detection by leveraging both visual and audio cues.}

\subsection{Audio-Visual Learning}
The integration of audio and visual information has emerged as a critical research direction in multimodal learning, as it not only enhances model performance but also facilitates a deeper understanding of complex scenes. Significant progress has been achieved in various aspects of audio-visual fusion~\cite{li2022learning, wei2022learning}. In audio-visual segmentation, researchers aim to accurately segment sound-producing objects based on audio-visual cues. Chen et al.~\cite{chen2024unraveling} proposed a novel informative sample mining method for audio-visual supervised contrastive learning. Ma et al.~\cite{ma2024stepping} introduced a two-stage training strategy to address the audio-visual semantic segmentation (AVSS) task. Building on these works, Guo et al.~\cite{guo2024open} introduced a new task: Open-Vocabulary AVSS (OV-AVSS), which extends AVSS to open-world scenarios beyond predefined annotation labels. Audio-visual event localization aims to identify the spatial and temporal locations of both visual and auditory events, with attention mechanisms widely used for modality fusion. For instance, He et al.~\cite{he2024cace} proposed an audio-visual co-guided attention mechanism, while Xu et al.~\cite{xu2020cross} introduced an audio-guided spatial-channel attention mechanism. Related tasks include audio-visual video parsing~\cite{zhou2024advancing, zhou2024label} and audio-visual action recognition~\cite{chalk2024tim}. Audio-visual anomaly detection~\cite{wu2020not, liu2024dual} has also become a growing research hot. For example, Yu et al.~\cite{yu2022modality} applied a self-distillation module to transfer single-modal visual knowledge to an audio-visual model, reducing noise and bridging the semantic gap between single-modal and multimodal features. Similarly, Pang et al.~\cite{pang2021violence} proposed a weighted feature generation approach, leveraging mutual guidance between visual and auditory information, followed by bilinear pooling for effective feature integration.

\subsection{Large Models in Video Understanding}
{In recent years, large models have exhibited exceptional capabilities in perception and reasoning for video understanding tasks, significantly accelerating the shift from purely visual models to multimodal video understanding frameworks. Representative visual models, such as VideoMAE~\cite{tong2022videomae}, employ masked self-supervised learning to effectively model spatiotemporal dynamics in videos, facilitating their widespread application in video classification, action recognition, and anomaly detection. With the success of VLMs like CLIP~\cite{radford2021learning} and ALIGN~\cite{jia2021scaling}, integrating language priors into video understanding has emerged as a prominent research trend. These models perform cross-modal semantic alignment through joint image-text encoding and have been widely adopted in tasks such as zero-shot action recognition, video retrieval, and open-vocabulary scene understanding. Further advances, including X-CLIP~\cite{ni2022expanding} and VideoCLIP~\cite{xu2021videoclip}, introduce temporal modeling into VLM architectures, significantly improving semantic comprehension of long-form video content. Meanwhile, VLM-based video reasoning tasks are gaining increasing attention. Models such as VL-T5~\cite{cho2021unifying} and VideoChat~\cite{li2023videochat} leverage language-guided mechanisms to enable video question answering, event interpretation, and causal reasoning, thereby substantially broadening the scope of video understanding.}

\section{Methodology}
\label{sec:method}

\begin{figure*}[t]
  \centering
  \includegraphics[width=0.99\linewidth]{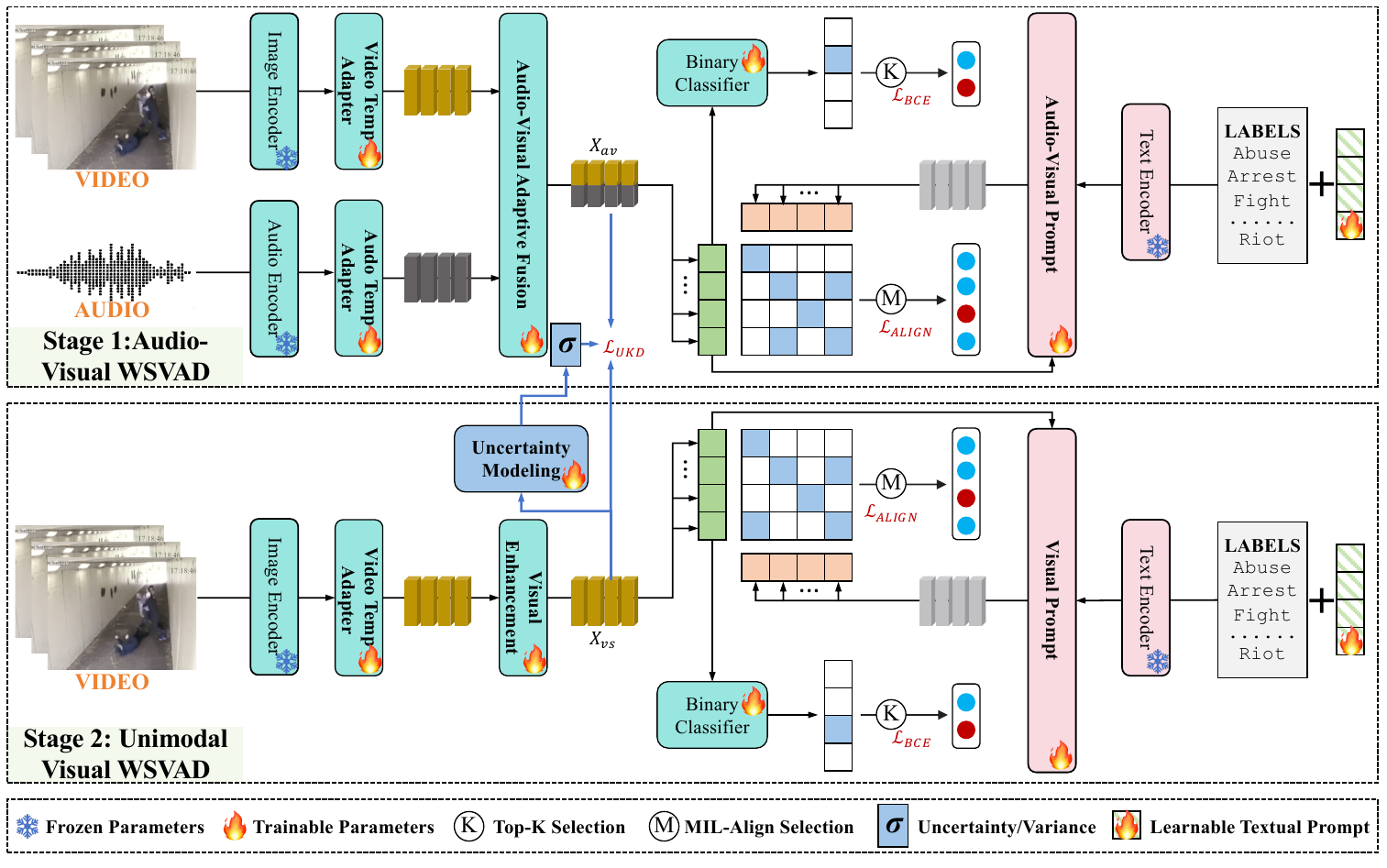}
  \caption{The pipeline of our proposed AVadCLIP. Our method supports both multimodal inputs and visual-only inputs via distillation, enabling robust video anomaly detection through the proposed UKD strategy. Throughout the entire framework, the pre-trained CLIP backbone remains fully frozen, with only a few modules being trainable. This design allows for efficient and lightweight adaptation of CLIP’s knowledge to the specific task of audio-visual anomaly detection.}
  \label{pipeline}
\end{figure*}

\begin{figure*}[t]
  \centering
  \includegraphics[width=1.0\linewidth]{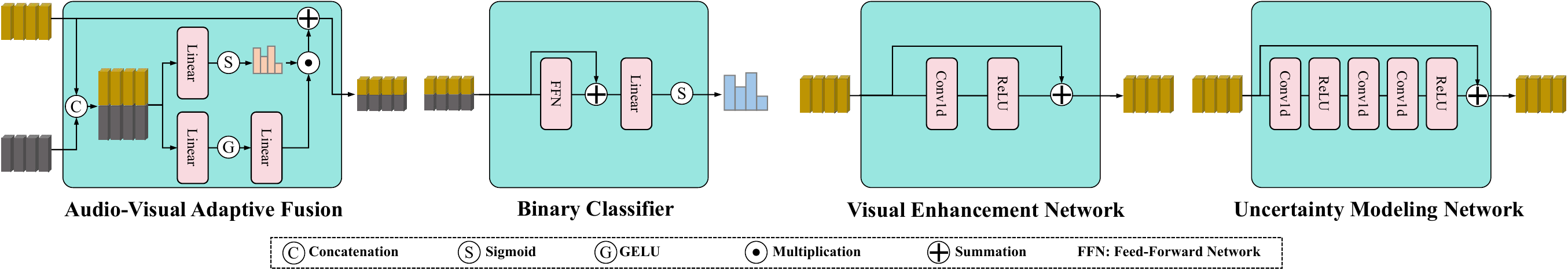}
  \caption{The pipeline of our proposed adaptive fusion module, binary classifier, visual enhancement network, and uncertainly modeling network.}
  \label{sub-pipeline}
\end{figure*}

\subsection{Problem Statement}
Given a training set of videos \(\{V_i\}\), where each video \(V\) contains both visual and corresponding audio information, along with a video-level label \(y \in \mathbb{R}^{C}\). Here, $C$ indicates that the number of categories (including the normal class and various anomaly classes). To facilitate model processing, we employ a video encoder and an audio encoder to extract high-level features \(X_v \in \mathbb{R}^{N \times d}\) and \(X_a \in \mathbb{R}^{N \times d}\), respectively, where \(N\) represents the temporal length of the video (i.e., the number of frames or snippets) and \(d\) denotes the feature dimensionality. The objective of WSVAD task is to train a detector using all available \(X_v\), \(X_a\), and their corresponding labels from the training set, enabling the model to accurately determine whether each frame in a test sample is anomalous and to identify the specific anomaly category.

The overall pipeline of our method as shown in Figure~\ref{pipeline}, starts with extracting features from video and audio using dedicated encoders, then adaptively fuses them for multimodal correspondence learning. We combine a classification branch with a CLIP-based alignment approach, using a audio-visual prompt to inject fine-grained multimodal information into text embeddings. Additionally, an uncertainty-driven distillation is employed to improve anomaly detection robustness in scenarios with incomplete modalities.

\subsection{Video and Audio Encoders}
\noindent \textbf{Video encoder.} 
Leveraging CLIP’s robust cross-modal representation, we use its image encoder (ViT-B/16) as a video encoder, in contrast to traditional models like C3D and I3D, which are less effective in capturing semantic relationships. We extract features from sampled video frames using CLIP, but to address CLIP’s lack of temporal modeling, we incorporate a lightweight temporal model, such as Graph Convolution Network (GCN)~\cite{zhong2019graph} and Temporal Transformer~\cite{wu2024vadclip}, to capture temporal dependencies. This approach ensures efficient transfer of CLIP’s pre-trained knowledge to the WSVAD task. 

\noindent \textbf{Audio encoder.} 
For audio feature extraction, we use Wav2CLIP~\cite{wu2022wav2clip}, a CLIP-based model that maps audio signals into the same semantic space as images and text. The audio is first converted into spectrograms, then sampled to match the number of video frames. These audio segments are processed by Wav2CLIP to extract features. To capture contextual relationships, we apply a temporal convolution layer~\cite{lea2016temporal}, which models local temporal dependencies, preserving key dynamics within the audio modality.

\subsection{Audio-Visual Adaptive Fusion }
In multimodal feature fusion, while both video and audio contain valuable semantic information, their importance often varies depending on the specific task. Inspired by human perception mechanisms~\cite{chen2024whole}, our approach follows a vision-centric, audio-assisted paradigm, where video features serve as the primary modality, and audio features complement and enhance visual information. To preserve the generalization capability of the original CLIP model in downstream tasks while avoiding the introduction of excessive trainable parameters, we design a lightweight adaptive fusion that integrates audio features without significantly increasing computational overhead. 
We present the structure of this fusion in Figure~\ref{sub-pipeline}.

Specifically, given the video feature $X_v$ and audio feature $X_a$, we first concatenate them to obtain a joint representation $X_{a+v}\in\mathbb{R}^{N \times 2d}$, which is then processed by two projection networks to generate the adaptive weight and residual feature. The first projection network computes adaptive fusion weights $W$, which determine the contribution of audio at each time step~\cite{chen2018pixelsnail}. This is achieved through a linear transformation followed by a sigmoid activation:
\begin{equation}
W = \text{Sigmoid}(\text{Linear}(X_{a+v})) \in \mathbb{R}^{N \times d}
\end{equation}

The second projection network is responsible for residual mapping, which transforms $X_{a+v}$ into a residual feature $X_{\text{res}}$ that encodes the fused information from both modalities:
\begin{equation}
X_{\text{res}} = \text{Linear}(\text{GELU}(\text{Linear}(X_{a+v}))) \in \mathbb{R}^{N \times d}
\end{equation}

Finally, the fused representation $X_{av}$ is obtained by adaptively incorporating the residual feature into the original video feature:
\begin{equation}
X_{av} = X_v + W \odot X_{\text{res}}
\end{equation}
where $\odot$  denotes element-wise multiplication. The adaptive weight $W$ dynamically adjusts the degree of audio integration, ensuring that video features remain dominant while audio features provide auxiliary information. Additionally, the residual mapping enhances the expressiveness of the fused representation by capturing nonlinear transformations. By introducing an adaptive fusion mechanism and maintaining a lightweight design, our fusion approach effectively balances efficiency and expressiveness, leveraging the complementarity of visual and audio modalities while minimizing computational overhead.

\subsection{Dual Branch Framework with Prompts}
We leverage a dual-branch framework~\cite{wu2024vadclip} for the WSVAD task, consisting of a classification branch and an alignment branch, which effectively leverage audio-visual information to improve detection accuracy. 
Classification branch consists of a lightweight binary classifier (as shown in Figure~\ref{sub-pipeline}), which takes $X_{av}$ as input and directly predicts the frame-level anomaly confidence $A$. 
Alignment branch leverages the cross-modal semantic alignment mechanism, which computes the similarity between frame-level features and class label features. To obtain class label representations, we leverage the CLIP text encoder combined with the learnable textual prompt~\cite{zhou2022learning} and audio-visual prompt to extract class embeddings, ensuring unified semantic alignment between visual and textual modalities. Given a set of predefined class labels (e.g., ``normal'', ``fighting''), we first introduce a learnable textual prompt, then we concatenate the textual prompt with class labels and feed them into CLIP text encoder to obtain the class representation $X_c$. 
Compared to the manually defined prompt, the learnable prompt allows the model to dynamically adjust textual representations during training, making them more suitable for the specific requirements of WSVAD. Furthermore, we incorporate an audio-visual prompt into the class label features to enrich the class representations with additional multimodal information. 

The proposed audio-visual prompt mechanism aims to dynamically inject instance-level key audio-visual information into text labels to enhance the representation. Specifically, we leverage the anomaly confidence $A$ from the classification branch and audio-visual features $X_{av}$ to generate a video-level global representation:
\begin{equation}
X_{p} = \text{Norm}\left(A^{\top}X_{av}\right) \in \mathbb{R}^{d}
\label{con:attention}
\end{equation}
where Norm represents the normalization operation. Next, we calculate the similarity matrix $S_p$ between the class representation $X_c$ and the global representation $X_{p}$ to measure the alignment between class labels and videos:
\begin{equation}
S_p = \text{Softmax}\left(X_c X_{p}^{\top}\right)
\end{equation}
based on $S_p$, we generate the enhanced instance-level audio-visual prompt $X_{mp}$:
\begin{equation}
X_{mp} = S_p X_{av} \in \mathbb{R}^{d}
\end{equation}
This operation dynamically adjusts the class representation’s focus on different video instances by calculating the similarity between global audio-visual features and class labels, thereby enhancing cross-modal alignment.

Then, we add $X_{mp}$ and the class representation $X_c$, followed by a feed-forward network (FFN) transformation and a skip connection to obtain the final instance-specific class embedding $X_{cp}$:
\begin{equation}
X_{cp} = \text{FFN}\left(\text{ADD}\left(X_{mp}, X_{c}\right)\right)+X_{c}
\label{con:visual_prompt}
\end{equation}
where ADD represents element-wise addition. 

This dual-branch framework provides anomaly confidence through the classification branch and refines category identification with class information via the alignment branch, improving robustness and enabling fine-grained anomaly detection.

\subsection{Optimization of Audio-Visual Model}
For the classification branch, we adopt the Top-K mechanism, as proposed in previous work~\cite{wu2020not}, to select the top $K$ anomaly confidence values from both normal and abnormal videos, which are averaged as the video-level prediction. The classification loss $\mathcal{L}_{BCE}$ is then computed using binary cross-entropy between the prediction and groundtruth class.

In the case of the alignment branch, the MIL-Align mechanism~\cite{wu2024vadclip} is applied. We compute an alignment map $M$, reflecting the similarity between frame-level features $X_{av}$ and all category embeddings $X_{cp}$. For each row in $M$, the top $K$ similarities are selected and their average is used to quantify the alignment between the video and the current class. This results in a vector $S = \{s_1, \ldots, s_m\}$ representing the similarity between the video and all possible classes. Then the multi-class prediction is then calculated as:
\begin{equation}
  p_i = \frac{exp\left(s_i/\tau\right)}{\sum_j{exp\left(s_j/\tau\right)}}
  \label{con:nce}
\end{equation}
where $p_i$ represents the prediction for the $i^{th}$ class, and $\tau$ is the temperature scaling parameter. Then, we compute $\mathcal{L}_{NCE}$ based on cross-entropy. Besides, to address the class imbalance in WSVAD , where normal samples dominate and anomaly instances are sparse, we employ the focal loss~\cite{lin2017focal}. Finally, the overall loss $\mathcal{L}_{ALIGN}$ for alignment branch is the average of $\mathcal{L}_{NCE}$ and $\mathcal{L}_{FOCAL}$.

\subsection{Uncertainty-Driven Distillation}

In the WSVAD task, audio serves as a complementary modality to video, enhancing detection accuracy. However, audio may be unavailable in practical scenarios, leading to performance degradation. To address this, we apply knowledge distillation by using a pre-trained multi-modal (video+audio) teacher model to guide a unimodal (video-only) student model, ensuring robust anomaly detection even without audio. 
Traditional knowledge distillation methods typically assume a deterministic transfer of knowledge, employing mean square error (MSE) loss to align the student model with the teacher’s feature representations. However, this approach fails to account for the inherent uncertainty in audio-visual feature fusion. In real-world scenarios, factors such as noisy audio or occluded visual content can introduce distortions in the fused features, leading to inaccurate feature representations and diminished generalization capability. 

To overcome this, we propose a probabilistic uncertainty distillation strategy~\cite{chang2020data, yang2022self}, which models data uncertainty during distillation, improving the student model’s robustness across diverse scenarios. Specifically, assume $X_{av,i} = X_{vs,i} + \epsilon\sigma_i$, where $\epsilon\sim \mathcal{N}(\mathbf{0},\mathbf{I})$, $X_{vs}$ represents enhanced visual features generated from the student model, and it is derived from $X_v$ after passing through a visual enhancement network, which is illustrated in Figure~\ref{sub-pipeline}. Besides, $\sigma_i$ refers to the inherent uncertainty between the $i^{th}$ pair of features. Then we model the observation as a Gaussian likelihood function to more accurately quantify data uncertainty in the feature distillation. The relationship between the audio-visual fusion feature $X_{av,i}$ and the unimodal feature $X_{vs,i}$ is formulated as:
\begin{equation}
p(X_{av,i} | X_{vs,i}, {\theta}) = \frac{1}{\sqrt{2\pi {\sigma}_i^2}} \exp\left(-\frac{\| X_{av,i} - X_{vs,i} \|^2}{2{\sigma}_i^2}\right)
\label{eq:pi}
\end{equation}
where $\theta$ is the parameter of models, to maximize the likelihood for each pair of features $X_{av,i}$ and $X_{vs,i}$, we adopt the log-likelihood form:
\begin{equation}
\ln p(X_{av,i} | X_{vs,i}, {\theta}) = -\frac{\| X_{av,i} - X_{vs,i} \|^2}{2{\sigma}_i^2} - \frac{\ln 2\pi {\sigma}_i^2}{2}
\label{eq:log_pi}
\end{equation}

In practice, we design a network branch (a simple three-layer convolutional neural network, which is shown in Figure 3) to predict the variance ${\sigma}_i^2$ and reformulate the likelihood maximization problem as the minimization of a loss function. Specifically, we employ an uncertainty-weighted MSE loss:
\begin{equation}
\mathcal{L}_{UKD} = \frac{1}{L} \sum_{i=1}^{L} \left[ \frac{\| X_{av,i} - X_{vs,i} \|^2 }{\sigma_{i}^2}+ \ln {\sigma}_i^2 \right]
\label{eq:loss_pi}
\end{equation}
where $L$ represents the number of feature pairs, and the constant term is omitted for clarity.

During the distillation process, the student model not only learns the unimodal feature $X_{vs,i}$ from the teacher model but also considers the feature uncertainty ${\sigma}_i^2$ to optimize its learning strategy. Specifically, the first term of the loss function represents the feature similarity between the student and teacher models, normalized by ${\sigma}_i^2$. This assigns smaller weights to features with higher uncertainty, thereby avoiding overfitting to hard-to-learn information. The second term acts as a regularization term to prevent ${\sigma}_i^2$ from becoming too small, ensuring effective distillation.

Ultimately, during the inference phase, we only input video and perform anomaly detection through the unimodal student model for audio-missing scenarios.

\section{Experiments}
\label{sec:experiements}

\subsection{Datasets and Evaluation Metrics}
\subsubsection{Datasets} 
We conduct extensive experiments on two audio-visual benchmarks: XD-Violence~\cite{wu2020not} and CCTV-Fights$_{sub}$~\cite{meng2025audio}, both of which contain synchronized audio and visual modalities. Unlike traditional unimodal datasets, these benchmarks enable a more comprehensive evaluation of our framework’s robustness under multimodal settings.~\textbf{XD-Violence.} As the largest publicly available audio-visual WSVAD dataset, XD-Violence~\cite{wu2020not} significantly surpasses existing datasets in scale and diversity. It comprises 3,954 training videos and 800 test videos, with the test set containing 500 violent and 300 non-violent videos. The dataset covers six distinct categories of violent events, including abuse, car accident, explosion, fighting, riot, and shooting, which occur at various temporal locations within videos. 

\noindent\textbf{CCTV-Fights$_{sub}$.} Derived from CCTV-Fights~\cite{perez2019detection}, CCTV-Fights$_{sub}$~\cite{meng2025audio} is a carefully curated subset designed to address audio-visual anomaly detection. The subset retains 644 high-quality videos depicting real-world fight scenarios, each with meaningful audio content, making it a valuable resource for evaluating audio-visual anomaly detection methods in real-world surveillance contexts.

\subsubsection{Evaluation Metrics}
For performance evaluation, we adopt distinct metrics tailored to different granularities of WSVAD tasks. For coarse-grained WSVAD, we employ frame-level Average Precision (AP), which provides a comprehensive measure of detection accuracy across varying confidence thresholds. For fine-grained anomaly detection, we utilize mean Average Precision (mAP)~\cite{wu2022weakly} computed across multiple intersection over union (IoU) thresholds and the average mAP (AVG) across different thresholds. Specifically, we evaluate mAP at IoU thresholds ranging from 0.1 to 0.5 with an interval of 0.1, followed by reporting AVG across these thresholds. 
\subsection{Implementation Details}
We conduct experiments on an NVIDIA RTX 4090 GPU, where the visual enhancement network is a single-layer 1D convolutional network, which includes a convolutional layer with a kernel size of 3, padding size of 1, ReLU activation function, and a skip connection. Such an operation effectively facilitates the aggregation of local contextual information. For input processing, we employ a frame selection strategy tailored to different datasets, sampling one frame per 16 frames for XD-Violence and one frame per 4 frames for CCTV-Fights$_{sub}$, using a uniform sampling strategy with a maximum frame count of 256; During optimization, we set the batch size, learning rate, and total epoch to 96, $1e^{-5}$, and 10, respectively.

\begin{table}[!t]
  \small
  \centering
  \caption{Coarse-grained comparisons on XD-Violence.}
  \setlength{\tabcolsep}{2.8pt}
  \begin{tabular}{lrcc}
    \toprule
    Method&Reference&Modality&AP(\%)\\
    \midrule
    DeepMIL~\cite{sultani2018real}&CVPR 2018&RGB(ViT)& 75.18\\
    Wu et al.~\cite{wu2020not} &ECCV 2020&RGB(ViT)&80.00\\ 
    RTFM~\cite{tian2021weakly} &ICCV 2021&RGB(ViT)&78.27\\ 
    AVVD~\cite{wu2022weakly}&TMM 2022&RGB(ViT)&78.10\\ 
    Ju et al.~\cite{ju2022prompting}&ECCV 2022&RGB(ViT)& 76.57\\
    DMU~\cite{zhou2023dual} &AAAI 2023&RGB(ViT)& 82.41\\
    CLIP-TSA~\cite{joo2023clip} &ICIP 2023&RGB(ViT)& 82.17\\
    AnomalyCLIP~\cite{zanella2024delving} & CVIU 2024& RGB(ViT) & 78.51 \\
    TPWNG~\cite{yang2024text}& CVPR 2024& RGB(ViT)& 83.68\\
    {VadCLIP~\cite{wu2024vadclip}} &AAAI 2024&RGB(ViT)& 84.51 \\
    \rowcolor{gray!20} \textbf{AVadCLIP$^*$} & \textbf{this work} & \textbf{RGB(ViT)} & \textbf{85.53} \\
    \midrule
    FVAI~\cite{pang2021violence} &ICASSP 2021&RGB(I3D)+Audio& 81.69\\
    MACIL-SD~\cite{yu2022modality} &ACMMM 2022&RGB(I3D)+Audio& 81.21\\
    CUPL~\cite{zhang2023exploiting} &CVPR 2023&RGB(I3D)+Audio& 81.43\\
    AVCL~\cite{meng2025audio} &TMM 2025&RGB(I3D)+Audio& 81.11\\
    \rowcolor{gray!20} \textbf{AVadCLIP} & \textbf{this work} & \textbf{RGB(ViT)+Audio} & \textbf{86.04} \\
  \bottomrule
\end{tabular}
\label{tabxd}
\end{table}

\begin{table}[!t]
  \small
  \centering
  \caption{Coarse-grained comparisons on CCTV-Fights$_{sub}$.}
  \setlength{\tabcolsep}{2.8pt}
  \begin{tabular}{lrcc}
    \toprule
    Method&Reference&Modality&AP(\%)\\
    \midrule
    VadCLIP~\cite{wu2024vadclip} &AAAI 2024&RGB(ViT)& 72.78 \\
    \rowcolor{gray!20} \textbf{AVadCLIP$^*$} & \textbf{this work} & \textbf{RGB(ViT)} & \textbf{73.36} \\
    \midrule
    MACIL-SD~\cite{yu2022modality} &ACMMM2022&RGB(I3D)+Audio& 72.92\\
    DMU~\cite{zhou2023dual} &CVPR2023&RGB(I3D)+Audio& 72.97\\
    AVCL~\cite{meng2025audio} &TMM 2025&RGB(I3D)+Audio& 73.20\\
    \rowcolor{gray!20} \textbf{AVadCLIP} & \textbf{this work} & \textbf{RGB(ViT)+Audio} & \textbf{73.38} \\
  \bottomrule
\end{tabular}
\label{tabfight}
\end{table}

\subsection{Comparison with State-of-the-Art Methods}

\begin{table*}[t]
  \centering
  \caption{Fine-grained comparisons on XD-Violence.}
  \begin{tabular}{lrccccccc}
    \toprule
    \multirow{2}{*}{Method} & \multirow{2}{*}{Reference} & \multirow{2}{*}{Modality} & \multicolumn{6}{c}{mAP@IoU(\%)}\\
    \cmidrule(lr){4-9}
    ~ & & & 0.1 & 0.2 & 0.3 & 0.4 & 0.5 & AVG  \\
    \hline
    Random & \multicolumn{1}{c}{-} & RGB(VIT) & 1.82 & 0.92 & 0.48 & 0.23 & 0.09 & 0.71  \\
    DeepMIL~\cite{sultani2018real} & CVPR 2018 & RGB(ViT) & 22.72 & 15.57 & 9.98 & 6.20 & 3.78& 11.65 \\
    AVVD~\cite{wu2022weakly} & TMM 2022 & RGB(ViT) & 30.51 & 25.75 & 20.18 &14.83 & 9.79& 20.21  \\
    {VadCLIP~\cite{wu2024vadclip}} & AAAI 2024 & RGB(ViT) & {37.03} & {30.84} & {23.38} & {17.90} & {14.31} & {24.70} \\
    \rowcolor{gray!20} \textbf{AVadCLIP$^*$} & \textbf{this work} & \textbf{RGB(ViT)} & \textbf{39.63} & \textbf{32.77} & \textbf{26.84} & \textbf{21.58} & \textbf{16.39} & \textbf{27.44} \\
    \rowcolor{gray!20} \textbf{AVadCLIP} & \textbf{this work} & \textbf{RGB(ViT)+Audio} & \textbf{41.89} & \textbf{34.61} & \textbf{27.08} & \textbf{22.16} & \textbf{17.30} & \textbf{28.61} \\
    \bottomrule
  \end{tabular}
  \label{tabfine-xd}
  \vspace{-0.2cm}
\end{table*}

\begin{table*}[!t]
  \centering
  \caption{Fine-grained comparisons on CCTV-Fights$_{sub}$.}
  \begin{tabular}{lrccccccc}
    \toprule
    \multirow{2}{*}{Method}  & \multirow{2}{*}{Reference} & \multirow{2}{*}{Modality} & \multicolumn{6}{c}{mAP@IoU(\%)}\\
    \cmidrule(lr){4-9}
    ~ & & & 0.1 & 0.2 & 0.3 & 0.4 & 0.5 & AVG  \\ \hline
    {VadCLIP~\cite{wu2024vadclip}} & AAAI 2024 & RGB(ViT) & {19.34} & {14.32} & {9.25} & {6.64} & {3.73} & {10.66} \\
    \rowcolor{gray!20} \textbf{AVadCLIP$^*$} & \textbf{this work} & \textbf{RGB(ViT)} &\textbf{21.10} & \textbf{14.57} & \textbf{9.01} & \textbf{5.74} & \textbf{4.69} & \textbf{11.02} \\
    \rowcolor{gray!20} \textbf{AVadCLIP} & \textbf{this work} & \textbf{RGB(ViT)+Audio} & \textbf{22.25} & \textbf{15.91} & \textbf{10.40} & \textbf{7.00} & \textbf{5.28} & \textbf{12.17} \\
    \bottomrule
  \end{tabular}
  \label{tabfine-cctv}
  \vspace{-0.2cm}
\end{table*}

\begin{table}[!t]
  \small
  \centering
  \caption{Cross-dataset WSVAD results on XD-Violence and CCTV-Fights$_{sub}$.}
    \begin{tabular}{c|cc}
    \toprule
    \qquad ~Test$\Rightarrow$& XD-Violence & CCTV-Fights$_{sub}$ \\
      Train$\Downarrow$ & AP(\%)  & AP(\%)  \\
    \hline
    XD-Violence & 86.04 & 69.24 \\ 
    CCTV-Fights$_{sub}$ & 76.60 & 73.38 \\ 
    \bottomrule
    \end{tabular}
   \label{tab-crossdata}
\end{table}%

\subsubsection{Performance comparison on XD-Violence}
Our experiments evaluate both coarse-grained and fine-grained anomaly detection performance on XD-Violence, comparing our AVadCLIP against state-of-the-art approaches, as shown in Tables~\ref{tabxd},~\ref{tabfine-xd}.

For coarse-grained anomaly detection, using only RGB inputs, AVadCLIP$^*$ ($*$ denotes RGB-only input) achieves an AP score of 85.53\%, surpassing all existing vision-only methods. Notably, it outperforms VadCLIP, the previous best-performing RGB-only approach, by 1.0\%, demonstrating superior visual anomaly detection. When incorporating audio, AVadCLIP further improves performance, significantly outperforming all multimodal baselines, achieving a remarkable 4.9\% gain over the latest method AVCL~\cite{meng2025audio}.

For fine-grained anomaly detection, AVadCLIP consistently outperforms all competitors across different IoU thresholds, as detailed in Table~\ref{tabfine-xd}. With RGB-only input, AVadCLIP$^*$ surpasses VadCLIP at all IoU thresholds, achieving an AVG improvement of 2.7\%. Similarly, the full-modality model AVadCLIP leads across all metrics, boosting the AVG by 3.9\%. These results highlight the effectiveness of multimodal learning in precisely localizing anomaly boundaries and improving category predictions. 

\subsubsection{Performance comparison on CCTV-Fights$_{sub}$}
The coarse-grained anomaly detection results on CCTV-Fights$_{sub}$ are presented in Table~\ref{tabfight}. For RGB-only methods, AVadCLIP$^*$ achieves 73.36\% AP, surpassing the state-of-the-art VadCLIP and demonstrating the effectiveness of our approach in unimodal scenarios. For audio-visual scenarios, AVadCLIP further improves performance, outperforming all existing methods. These results indicate that incorporating audio information can further enhance anomaly detection performance, validating the effectiveness of cross-modal complementary information mining.

We present the fine-grained anomaly detection results on CCTV-Fights$_{sub}$ in Table~\ref{tabfine-cctv}, and it can be observed that AVadCLIP consistently outperforms all competitors at different IoU thresholds. Using RGB-only input, AVadCLIP$^*$ surpasses VadCLIP at all IoU thresholds, achieving a 0.4\% improvement in AVG. Similarly, AVadCLIP with audio leads in all metrics, increasing AVG by 1.5\%. These results further highlight the effectiveness of multimodal learning in accurately locating anomalous boundaries and improving category prediction.

\subsubsection{Cross-dataset Results}
{Table~\ref{tab-crossdata} presents the cross-dataset evaluation results of AVadCLIP on XD-Violence and CCTV-Fights$_{sub}$, aiming to assess its generalization capability across different domains. Despite being trained on one dataset and tested on another, AVadCLIP consistently achieves competitive performance, demonstrating strong robustness and transferability. For example, AVadCLIP trained on XD-Violence still achieves an AP of 69.24\% when directly tested on the surveillance-oriented CCTV-Fights$_{sub}$, with less than a 4\% drop compared to the model trained specifically on that dataset. These results highlight the model’s ability to generalize well to unseen data distributions and diverse anomaly scenarios.

Overall, AVadCLIP achieves state-of-the-art performance in both unimodal and multimodal settings across coarse-grained and fine-grained anomaly detection tasks. The comprehensive results validate its effectiveness in leveraging audio-visual collaboration and demonstrate the feasibility of uncertainty-driven distillation strategy.}

\subsection{Ablation Studies}

\begin{table}[t]
  \small
  \centering
    \caption{Effectiveness of designed modules on XD-Violence.}
    \begin{tabular}{ccc|cc}
    \toprule
    AV Fusion & AV Prompt & $\mathcal{L}_{FOCAL}$  & \multicolumn{1}{c}{AP(\%)} & \multicolumn{1}{c}{AVG(\%)} \\
    \hline
    $\times$ & $\times$ & $\times$ & 79.85 & 27.89 \\
    $\surd$ & $\times$& $\times$ & 82.90 & 26.63 \\
    $\surd$ & $\surd$ &$\times$ & \textbf{86.18} & 26.79 \\
    $\surd$ & $\surd$ & $\surd$ & 86.04 & \textbf{28.61} \\
    \bottomrule
    \end{tabular}
    \label{tab-ablation}%
\end{table}%

\begin{table}[!t]
  \small
  \centering
  \caption{Effectiveness of audio-visual fusion  on XD-Violence.}
    \begin{tabular}{l|cc}
    \toprule
    Method & \multicolumn{1}{c}{AP(\%)} & \multicolumn{1}{c}{AVG(\%)} \\
    \hline
    Cross Attention & 75.15 & 10.51 \\
    Element-wise Addition & 83.02 & 27.66 \\
    Concat+Linear Projection & 83.36 & \textbf{28.88} \\
    Adaptive Fusion & \textbf{86.04}  & 28.61 \\
    \bottomrule
    \end{tabular}
     \label{tab-fusion}%
\end{table}%

\begin{table}[!t]
  \small
  \centering
  \caption{Effectiveness of UKD on XD-Violence.}
    \begin{tabular}{l|cc}
    \toprule
     {Method} & \multicolumn{1}{c}{AP(\%)} & \multicolumn{1}{c}{AVG(\%)} \\
    \hline
    Audio Model w/o UKD & 50.89 & 12.20 \\
    Audio Model w\quad UKD & \textbf{52.51}  & \textbf{13.50} \\
    \hline
    Visual Model w/o UKD & 84.60 & 22.92 \\
    Visual Model w\quad UKD & \textbf{85.53}  & \textbf{27.44} \\
    \hline
    Audio-Visual Model & \textbf{86.04} & \textbf{28.61} \\
    \bottomrule
    \end{tabular}
     \label{tab-ukd}%
\end{table}%

\begin{table}[!t]
  \small
  \centering
  \caption{Effectiveness of UKD on CCTV-Fights$_{sub}$.}
    \begin{tabular}{l|cc}
    \toprule
     {Method} & \multicolumn{1}{c}{AP(\%)} & \multicolumn{1}{c}{AVG(\%)} \\
    \hline
    Visual Model w/o UKD & 67.89 & 10.65 \\
    Visual Model w\quad UKD & \textbf{73.36}  & \textbf{11.02} \\
    \hline
    Audio-Visual Model & \textbf{73.38} & \textbf{12.17} \\
    \bottomrule
    \end{tabular}
     \label{tab-cctv-ukd}%
\end{table}%

\subsubsection{The effect of audio-visual adaptive fusion}
From Table~\ref{tab-ablation}, it can be observed that the introduction of audio-visual fusion improves detection performance. Furthermore, Table~\ref{tab-fusion} presents the impact of different audio-visual fusion strategies on anomaly detection performance. First, the cross attention fusion performs poorly in the WSVAD task, indicating that although it can capture the relationships between modalities, its complex parameterized design may negatively impact the generalization ability of CLIP model in downstream WSVAD tasks. Next, the simple element-wise addition strategy achieves an AP of 83.02\% and an AVG of 27.66\%. Then, the concatenation with linear projection approach improves the AP to 83.36\% and the AVG to 28.88\%, indicating that enhancing feature representation through linear transformation facilitates more effective cross-modal information capture. Finally, our proposed adaptive fusion strategy achieves the best AP of 86.04\%, outperforming the other three methods on the whole. This demonstrates that our adaptive fusion strategy, as a lightweight and effective fusion strategy, can more exploit complementary information between audio and visual modalities. 

\begin{figure*}[!t]
  \centering
  \includegraphics[width=0.98\linewidth]{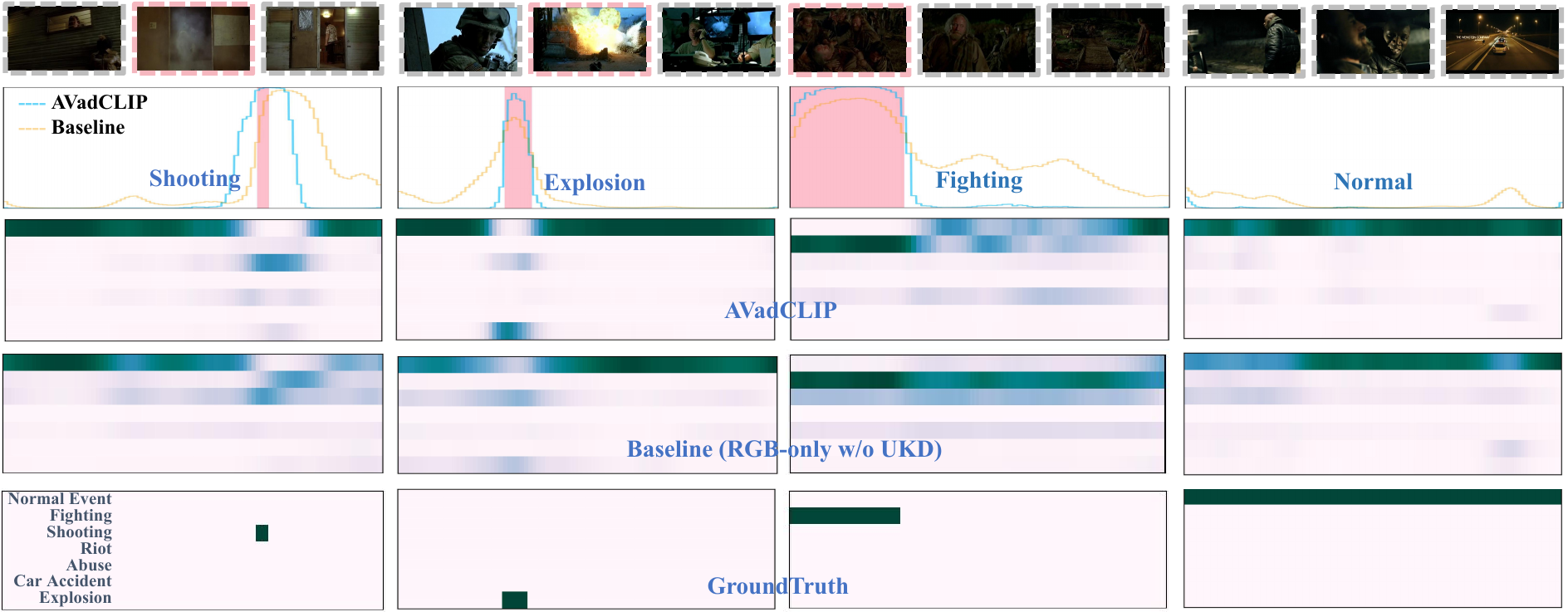}
  \caption{Coarse-grained and Fine-grained WSVAD visualization results of AVadCLIP and the baseline model on XD-Violence.}
  \label{vis}
\end{figure*}

\begin{figure*}[!t]
  \centering
  \includegraphics[width=1.0\linewidth]{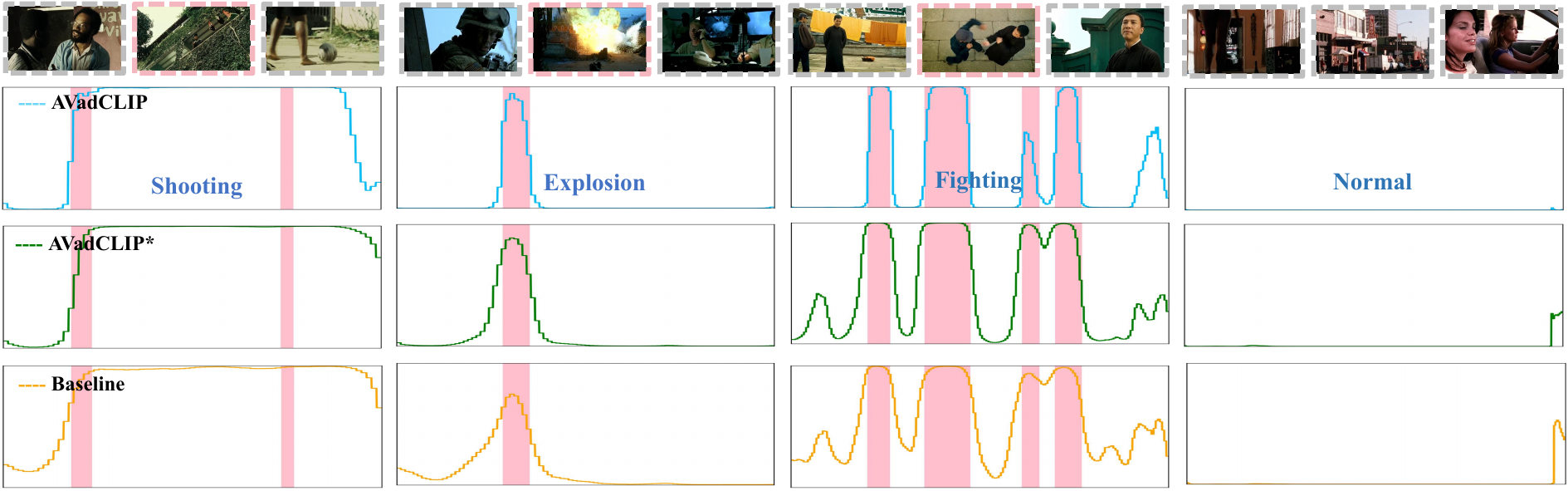}
  \caption{Coarse-grained WSVAD visualization results of AVadCLIP, AVadCLIP$^*$, and the baseline model on XD-Violence.}
  \label{vis-xd}
\end{figure*}

\begin{figure*}[!t]
  \centering
  \includegraphics[width=1.0\linewidth]{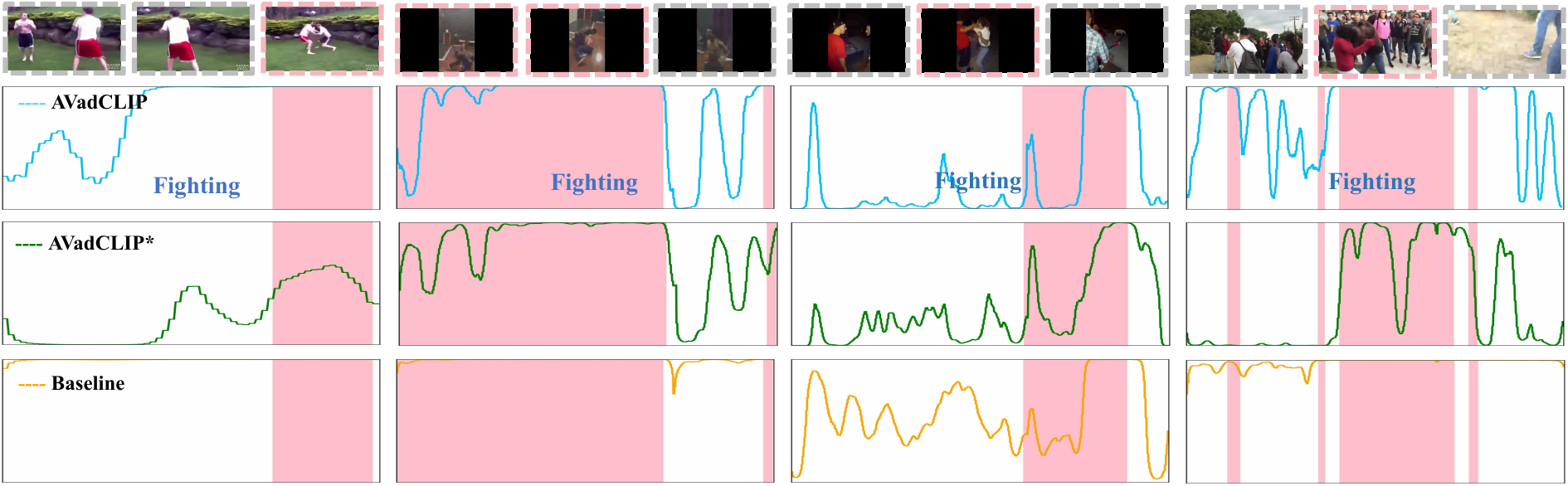}
  \caption{Coarse-grained WSVAD visualization results of AVadCLIP, AVadCLIP$^*$, and the baseline model on CCTV-Fights$_{sub}$.}
  \label{vis-cctv}
\end{figure*}

\subsubsection{The effect of audio-visual prompt and $\mathcal{L}_{FOCAL}$}
As presented in Table~\ref{tab-ablation}, the baseline model achieves an AP of only 79.85\%. Integrating the audio-visual prompt on top of the adaptive fusion mechanism significantly enhances performance, increasing the AP to 86.18\%. This improvement underscores the effectiveness of the audio-visual prompt in capturing critical multimodal patterns, thereby facilitating more precise anomaly recognition. Furthermore, incorporating focal loss into the model contributes to refining anomaly boundary detection, leading to more stable performance in fine-grained anomaly localization. In summary, the audio-visual prompt primarily enhances coarse-grained anomaly detection, and focal loss further refines boundary precision, enabling the model to achieve optimal performance across both AP and AVG metrics.

\subsubsection{The effect of uncertainty-driven distillation}
As shown in Table~\ref{tab-ukd}, the proposed UKD mechanism significantly enhances anomaly detection performance in both visual-only and audio-only models. Specifically, in the visual-only setting, UKD achieves a 0.9\% improvement in AP and a 4.5\% increase in AVG, attaining performance levels comparable to the teacher model trained with audio-visual inputs. Similarly, the audio-only model also benefits from UKD, exhibiting consistent performance gains. These results highlight the effectiveness of UKD in leveraging data uncertainty to enhance the robustness of unimodal representations during the distillation process, making it particularly well-suited for real-world applications where modality incompleteness is prevalent.
In addition, we present a comparison of the effectiveness of the UKD module on CCTV-Fights$_{sub}$ in Table~\ref{tab-cctv-ukd}. The results show that the proposed UKD mechanism significantly improves the anomaly detection performance of the unimodal model. Notably, in the visual-only scenario, adding UKD improves AP by 5.5\%, achieving performance comparable to that of the audio-visual model. This finding further demonstrates the effectiveness of UKD in enhancing the robustness of unimodal representations through data uncertainty.

\subsection{Qualitative Results}
In Figure~\ref{vis}, we present the qualitative visualizations of AVadCLIP and the baseline model for both coarse-grained and fine-grained WSVAD. The blue curves denote the anomaly predictions by AVadCLIP, whereas the yellow curves represent those by the baseline model (RGB-only w/o UKD). As illustrated, compared to the baseline model, AVadCLIP significantly reduces anomaly confidence in normal video segments, thereby enhancing its ability to distinguish between abnormal and normal regions more accurately. The fine-grained map below also indicates that AVadCLIP can predict categories with greater precision. Notably, the observed performance improvement supports our hypothesis that audio information is more advantageous in visual occlusion (shooting) or acoustic dominant scenes (explosion), and can effectively eliminate ambiguity in visually similar patterns in anomaly detection scenes, thereby ensuring more robust detection performance.

In addition, we compare the coarse-grained visualization results of the baseline model (RGB-only w/o UKD), the student model AVadCLIP$^*$, and the teacher model AVadCLIP on XD-Violence, as shown in Figure~\ref{vis-xd}. Experimental results show that AVadCLIP significantly outperforms the other two counterparts. By using this model as a teacher model to guide the unimodal student model, it effectively mitigates anomaly confidence biases, steering them towards more accurate detection results. To a certain extent, this demonstrates the robustness of our proposed method.

{In order to demonstrate the superiority of our proposed method more comprehensively and intuitively, Figure~\ref{vis-cctv} shows the coarse-grained visualization results on the CCTV-Fights$_{sub}$ dataset (since this dataset only includes the ``Fighting" category, fine-grained visualizations are not provided). It can be seen that our method achieves significantly higher anomaly confidence scores in abnormal regions and notably lower scores in normal regions compared to the baseline model. This demonstrates that the integration of audio and video information can still yield substantial performance improvements in complex scenes. Besides, as can be seen from the last two rows, the unimodal model distilled with the UKD mechanism shows significantly fewer false positives compared to the baseline, demonstrating that the UKD mechanism effectively transfers audio-visual multi-modal knowledge into the unimodal model.}

\section{Conclusion}
\label{sec:conclusion}
In this work, we propose a novel weakly supervised framework for robust video anomaly detection using audio-visual collaboration. Leveraging the powerful representation ability and cross-modal alignment capability of CLIP, we design two distinct modules to achieve efficient audio-visual collaboration and multimodal anomaly detection, based on the frozen CLIP model. Specifically, to seamlessly integrate audio-visual information, we introduce a lightweight fusion mechanism that adaptively generates fusion weights based on the importance of audio to assist visual information. Additionally, we propose an audio-visual prompt strategy that dynamically refines text embeddings with key multimodal features, strengthening the semantic alignment between video content and corresponding textual labels. To further bolster robustness in scenarios with missing modalities, we develop an uncertainty-driven distillation module that synthesizes audio-visual representations from visual inputs, focusing on challenging features. Experimental results across two benchmarks demonstrate that our framework effectively enables video-audio anomaly detection and enhances the model’s robustness in scenarios with incomplete modalities. In the future, we will explore the integration of additional modalities (e.g., textual description) based on VLMs to achieve more robust video anomaly detection.

\bibliographystyle{IEEEtran}
\bibliography{ref}

\end{document}